\documentclass[10pt,twocolumn,letterpaper]{article}

\usepackage{iccv}
\usepackage{times}
\usepackage{epsfig}
\usepackage{graphicx}
\usepackage{amsmath}
\usepackage{amssymb}
\usepackage{subcaption}
\usepackage[toc,page]{appendix}
\usepackage{placeins}

\usepackage[table]{xcolor}

\definecolor{red}{rgb}{0.95,0.4,0.4}
\definecolor{blue}{rgb}{0.4,0.4,0.95}

\DeclareMathOperator*{\maxf}{max}

\usepackage[breaklinks=true,bookmarks=false]{hyperref}

\iccvfinalcopy

\def\iccvPaperID{3674} 

\ificcvfinal\fi

\begin{document}

\title{Presence-Only Geographical Priors for Fine-Grained Image Classification}

\author{Oisin Mac Aodha\hspace{20pt}Elijah Cole\hspace{30pt}Pietro Perona\\Caltech \\\url{www.vision.caltech.edu/~macaodha/projects/geopriors}}


\maketitle
\ificcvfinal\thispagestyle{empty}\fi


%
%
%
\begin{abstract}
Appearance information alone is often not sufficient to accurately differentiate between fine-grained visual categories.
Human experts make use of additional cues such as where, and when, a given image was taken in order to inform their final decision.
This contextual information is readily available in many online image collections but has been underutilized by existing image classifiers that focus solely on making predictions based on the image contents.

We propose an efficient spatio-temporal prior, that when conditioned on a geographical location and time, estimates the probability that a given object category occurs at that location.
Our prior is trained from presence-only observation data and jointly models object categories, their spatio-temporal distributions, and photographer biases.
Experiments performed on multiple challenging image classification datasets show that combining our prior with the predictions from image classifiers results in a large improvement in final classification performance.
\end{abstract}

\section{Introduction}
\begin{figure}[t]
\begin{center}
  \includegraphics[width=\columnwidth]{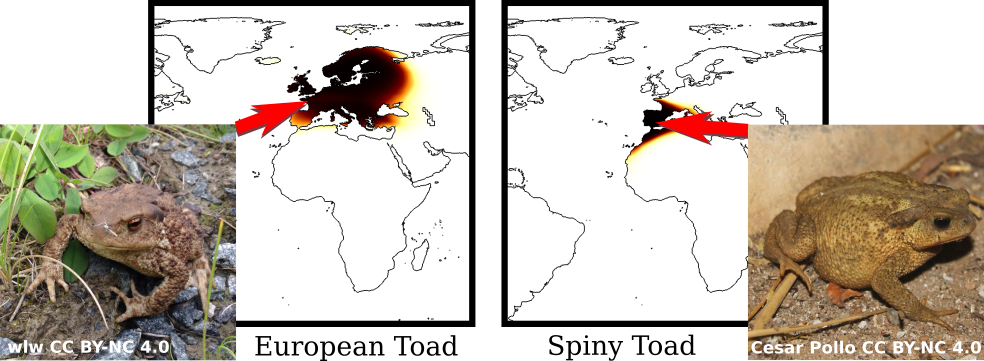}
\end{center}
  \vspace{-10pt}
  \caption{Differentiating between two visually similar categories such as the European (left) and Spiny (right) Toad can be challenging without additional context. To address this problem, we propose a spatio-temporal prior that encodes where, and when, a given category is likely to occur. For a known test location our prior predicts how likely it is for each category to be present. Darker colors indicate locations that are more likely to contain the object of interest.}
\label{fig:teaser}
\end{figure}

Correctly classifying objects into different fine-grained visual categories is a challenging problem.
In contrast to generic object recognition, it can require knowledge of subtle features that are essential for differentiating between visually similar categories.
However, without having access to additional information that may not be present in an image, many categories can be visually indistinguishable.
For example, the two toad species in Fig.~\ref{fig:teaser} are similar in appearance but tend to be found in very different locations in Europe.
Knowing \emph{where} a given image was taken can provide a strong prior for \emph{what} objects it may contain.

Most images that are captured and shared online today also come with additional metadata in the form of \emph{where} they were taken, \emph{when} they were taken, and \emph{who} captured them.
This information not only offers the possibility of helping to resolve ambiguous cases for image classification, but can also enable us to generate predictions of where, and when, different objects are likely to be observed.

Existing work that uses location information to improve classification performance either discretizes the input data into spatio-temporal volumes \cite{berg2014birdsnap}, store the entire training set in memory at inference time \cite{wittich2018recommending}, or jointly train deep images classifiers along with corresponding location information \cite{tang2015locationcontext}.
Methods that discretize or store the raw training data do not scale well in terms of memory, and jointly training image classifiers with location information necessitates that location information is present at test time - which may not always be the case.
We take inspiration from species distribution modeling (SDM) \cite{fink2010spatiotemporal}, and instead model a separate geographical prior that can be combined with the predictions of \emph{any} image classifier.
However, unlike many approaches to SDM that assume they have access to presence and absence information at training time (\eg \cite{tang2018VAE}), we make a more general assumption that only presence information is available \ie we know where the categories have been observed, but have \emph{no} explicit data regarding where they are not found.

In this work we make the following contributions:
(1) An efficient spatio-temporal prior that jointly models the relationship between location, time of year, photographer, and the presence of multiple different object categories.
(2) A novel presence-only training loss to capture these relationships.
(3) Experiments that show that combining the probabilistic predictions of image classifiers with our prior significantly improves the test time performance on challenging fine-grained image datasets.

\vspace{-5pt}
\section{Related Work}
\vspace{-5pt}
Here we discuss work related to spatio-temporal models that encode the location of a set of discrete object categories.
We do not address methods that explore other uses of location information such as inferring where an image was taken given only the raw pixels \cite{hays2008im2gps,vo2017revisiting}, or methods that use location to disambiguate visually similar places for image localization \cite{vishal2015accurate,zhai2018learning}.

%
\vspace{-5pt}
\subsubsection*{Fine-Grained Image Classification}
\vspace{-5pt}
Correctly determining which one of multiple possible fine-grained categories is present in an image requires understanding the relationship between subtle visual features and the corresponding image-level category label \eg \cite{wah2011caltech,khosla2011novel,yang2015large,van2018inaturalist}.
Existing approaches have investigated the modeling of parts \cite{liu2012dog,zhang2013deformable,branson2014bird,zhang2014part,huang2016part}, higher order feature interactions \cite{lin2015bilinear,gao2016compact}, attention mechanisms \cite{xiao2015application,zhao2017diversified,wang2017residual}, noisy web data \cite{krause2016unreasonable}, novel training losses \cite{cui2019class}, and pairwise category information
\cite{dubey2018pairwise}.
Orthogonal to those works, we propose a spatio-temporal prior that can be combined with the probabilistic predictions of any image classifier to improve the final classification performance.

%
\vspace{-5pt}
\subsubsection*{Location and Classification}
\vspace{-5pt}
A small number of approaches have explored the use of location information to improve image classification at test time.
Berg \etal~\cite{berg2014birdsnap} proposed a spatio-temporal prior that when combined with the output of an image classifier increased the accuracy of bird species classification.
Their approach discretized location and time into spatio-temporal cubes and used an adaptive kernel density estimator to represent the distribution of each species independently.
Also in the context of predicting the presence of different biological species, Wittich \etal~\cite{wittich2018recommending} evaluated different nearest neighbor based lookup strategies for retrieving the most relevant instances from a training set of geo-tagged observations.
These approaches are inefficient in terms their memory requirements as they necessitate storing either the entire training set or a discretized version of it.
Existing repositories of citizen science data (\eg \cite{sullivan2009ebird,iNatWeb,gbifWeb}) can contain on the order of tens of millions of observations making them prohibitively large to store and retrieve on mobile devices.
Choosing the correct discretization is challenging \cite{openshaw1983modifiable}, and incorrect choices can significantly affect the final performance \cite{lechner2012investigating,moat2018refining}. 
A key benefit of our approach is that discretization is not required.

Tang \etal\cite{tang2015locationcontext} explored different feature encodings for incorporating location information directly into deep neural networks at training time.
This included raw location features (\ie longitude and latitude), demographic information collected via a census, user provided hash-tags, and geographical map features (\eg land use estimates).
The disadvantage of their method is that it assumes that location information is present at test time and that all the required features can be computed for a given test location.
Furthermore, they cannot use location information that does not have an associated image. 
They also need to retrain their entire model if new location data is collected. 
We instead propose an efficient spatio-temporal prior that jointly models the spatial distribution of multiple object categories that can be trained independently of the image classifier. 
Parallel to our work, \cite{chu2019geo} builds on \cite{tang2015locationcontext} by exploring different ways to integrate location information into deep image classifiers. 


%
\vspace{-5pt}
\subsubsection*{Spatio-Temporal Distribution Modelling}
\vspace{-5pt}
Our goal is to estimate the spatio-temporal distribution of a set of object categories.
Related to this, there is a rich literature exploring models for estimating the distribution of biological specimens across geographic space and time~ \cite{hegel2010current}.
This is referred to as species distribution modelling or environmental niche modelling.
Broadly, these methods can be divided into two groups, those that use \emph{presence-absence} data and those that use \emph{presence-only} data~\cite{hastie2013POControversy}.

Making a presence-absence observation at a given location requires that every species from a predefined set of interest be confirmed as either present or absent for that sampling event.
In practice, this kind of data is onerous to collect because it requires intense survey effort to confirm that a species is absent with a high degree of certainty \cite{mackenzie2005issues}.
However, once this data is collected it can be combined with standard supervised classification approaches such as logistic regression \cite{hastie2013POControversy}, probit regression~\cite{pollock2014understanding}, Gaussian processes \cite{golding2016fast}, decision trees \cite{fink2010spatiotemporal}, and neural networks \cite{yen2004marbled,ozesmi1999artificial,mastrorillo2003fish}, among others \cite{elith2009dothey,norberg2019comprehensive}.
Presence-absence data is also compatible with traditional multi-label learning \cite{jones2011multi,boutell2004learning,zhang2006multi,chen2012feature,wang2016cnnrnn}.
Recently deep models have been applied to this problem in order to jointly model the location preferences of different species \cite{harris2015generating,chen2017DMSE,franceschini2018cascaded,tang2018VAE,botella2018deep} and human sampling biases \cite{chen2018bias}.

In contrast, a presence-only (\ie incidental) observation may be recorded wherever an object of interest is encountered - \emph{without} requiring any absences to be verified.
While presence-only data can be much easier to collect, the lack of absence information makes it more difficult to model.
This limitation is typically dealt with in one of three different ways.
The first approach is to generate `pseudo-negatives' and then apply one of the presence-absence approaches from above.
As no true negative information is available, these approaches randomly sample a set of locations and make the assumption that these locations are absences \eg \cite{engler2004improved,phillips2009sample,barbet2012pseudonegs}.
The second commonly used approach is to train a highly regularized model directly on the presence-only data \eg by fitting a maximum entropy distribution \cite{phillips2004maximum} or a low-rank model \cite{fithian2018flexible}, forcing the model to explain data where it has been observed and to be uncertain elsewhere.
Finally, and most related to our work, there are approaches that use additional information such as the detectability of a given species and a photographer's propensity to image them \eg
\cite{mackenzie2002estimating,garrard2013general}. 

Unlike many of the classic approaches for spatio-temporal distribution modelling, in this work we jointly learn a continuous spatio-temporal prior for each category of interest using a neural network to amortize the computation.
In contrast to previous deep distribution models \eg \cite{harris2015generating,chen2017DMSE,tang2018VAE}, we do not require presence-absence data or additional environmental features as input.
We instead exploit the structure that exists in online image repositories, such as those collected by citizen scientists, to jointly model objects, their locations, and photographer biases.

%
%





%
%
%
\section{Methods}
\vspace{-5pt}
Here we outline our spatio-temporal prior, which models the geographical and temporal distribution of a set of object categories and photographers. 
During training we assume that we have access to a set of tuples $\mathcal{D} = \{(I_i, \mathbf{x}_i, y_i, p_i) | i=1, ..., N\}$, where $I_i$ is an image, $y_i \in \{1, ...., C\}$ is the corresponding class label, $\mathbf{x}_i = [\mathrm{lon}_i, \mathrm{lat}_i, \mathrm{time}_i]$ represents the location (longitude and latitude) and time the image was taken, and $p_i$ is the individual, \ie photographer, who captured the image. 
Note that the location does not need to be captured alongside the image.
$\mathcal{D}$ can be assembled from unrelated image and location datasets as long as both contain the same categories. 

At test time, given an image and where and when it was taken we aim to estimate which category it contains \ie $P(y|I,\mathbf{x})$. 
One approach is to model the joint distribution $P(I,\mathbf{x})$ as in \cite{tang2015locationcontext}, but this necessitates that the location information is \emph{always} available at test time. 
Instead, inspired by \cite{berg2014birdsnap}, we can incorporate location information as a Bayesian spatio-temporal prior.
If we assume that $I$ and $\mathbf{x}$ are conditionally independent given $y$, then 
\begin{align}
    P(y|I,\mathbf{x}) &=& \frac{P(I,\mathbf{x}|y)P(y)}{P(I,\mathbf{x})}\\
    &=& \frac{P(I) P(\mathbf{x})}{P(I,\mathbf{x})} \frac{P(y|I) P(y|\mathbf{x})}{P(y)}\\
    & \propto & P(y|I)P(y|\mathbf{x}),
    \label{eqn:prob_model}
\end{align} 
where we assume a uniform prior $P(y) = 1 / C$ for $y\in\{1\ldots,C\}$. 
In reality an image may contain location information unrelated to the class label (\eg the background), but we assume this factorization is valid. 
By factoring the distribution in this way we can represent the image classifier, $P(y|I)$, and spatio-temporal prior, $P(y|\mathbf{x})$, separately.
Note that at test time we do not assume that we have any knowledge of the individual $p$ who captured the image.  
In this work we focus our attention on representing $P(y|\mathbf{x})$. 
For $P(y|I)$ we can use any discriminative model that produces a probabilistic output \eg a convolutional neural network.

\begin{figure}[t]
\begin{center}
  \includegraphics[width=0.9\columnwidth]{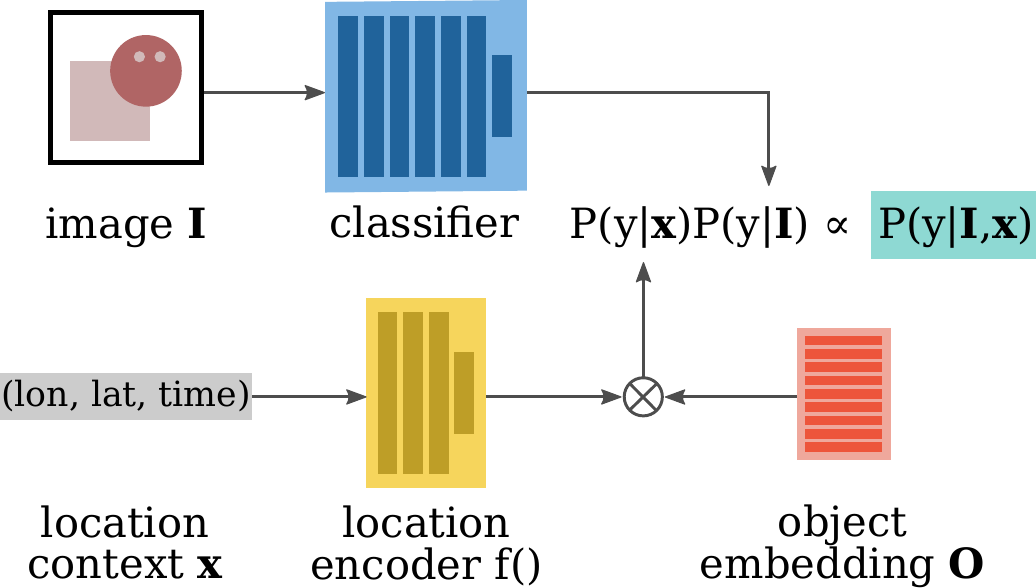}
\end{center}
  \vspace{-10pt}
  \caption{{\bf Inference time.} Our goal is to estimate if an object category $y$ is present in an input image $\mathbf{I}$. At test time we make use of additional spatio-temporal information $\mathbf{x}$ in the form of where and when the image was taken.} 
\label{fig:overview}
\end{figure}

\subsubsection*{Presence-Absence Loss}
As we are modeling the spatio-temporal prior independently from the image classifier our training data is now of the form $\mathcal{D} = \{(\mathbf{x}_i, y_i, p_i) | i=1, ..., N\}$. 
In the ideal case we would have complete information consisting of where and when a given category has both been observed to be present and observed to be \emph{not} present \eg as in \cite{chen2017DMSE,tang2018VAE}.
Then instead of $y_i \in {1, ..., C}$, each spatio-temporal location $\mathbf{x}_i$ would be associated with a binary multi-label vector $\mathbf{y}_i = [y_i^1, ..., y_i^C]$ where each entry $y_i^c \in \{0,1\}$ indicates whether or not category $c$ has been observed as being present at $\mathbf{x}_i$.
This formulation results in a standard multi-label learning problem, enabling us to estimate the parameters of the spatio-temporal model by solving
\begin{equation}
    \max_\theta \sum_{i=1}^{N}\sum_{c=1}^{C} y_i^c \log(\hat{y}_i^c) + (1-y_i^c) \log(1-\hat{y}_i^c),
    \label{eqn:pres_abs}
\end{equation}
where we define $\hat{y}_i^c = P(y_i^c | \mathbf{x}_i)$ and $P$ is parameterized by $\theta$.
However, as discussed previously, presence-absence information is both difficult and time consuming to acquire in real world settings.

\subsubsection*{Presence-Only Loss}
In this work we explore the more challenging presence-only setting where each spatio-temporal location $\mathbf{x}_i$ is associated with a single label $y_i \in \{1, ...., C\}$ indicating which category was observed.
In essence, we have a label vector $\mathbf{y}_i$ where there is only one affirmative entry, \ie $y_i^c = 1$ for some $c$, and the remaining entries are unknown. 
In this setting, Eqn. \ref{eqn:pres_abs} can be written as 
\begin{equation}
    \maxf_\theta \sum_{i=1}^{N} \log(\hat{y}_i^{c_i}) + A_i,
    \label{eqn:pres_only}
\end{equation}
where $A_i$ represents a proxy absence term for the $i^{th}$ training example and $c_i$ is the corresponding observed category.
Now the question becomes how to choose $A_i$.

One common approach for representing $A_i$ is to generate `pseudo-negatives' \cite{barbet2012pseudonegs} by randomly sampling absence data from some parametric distribution.
For instance, one might set
\begin{equation}
    A_{i} = \log(1-P(y_i|\mathbf{r}_i)).
\label{eqn:rand_prior}
\end{equation}
where $\mathbf{r}_i$ is a randomly selected spatio-temporal location with $[\mathrm{lon}(\mathbf{r}_{i}),\mathrm{lat}(\mathbf{r}_i)] \sim \mathrm{Unif}(\mathbb{S}^2)$ and $\mathrm{time}(\mathbf{r}_i) \sim \mathrm{Unif}([0,1])$.
The implicit assumption is that each category (whether man-made or naturally occurring) occurs in a relatively small subset of $\mathbb{S}^2 \times [0,1]$, so the probability of a category occurring at a randomly chosen location $\mathbf{r} \in \mathbb{S}^2 \times [0,1]$ is small as well. 
To the extent that this assumption holds, these pseudo-negatives are likely to be valid. 

An alternative approach is to instead sample absences over locations and times where the presence data for other categories occurs. 
In this case we would set $A_i$ according to Eqn. \ref{eqn:rand_prior} but sample negative locations from the positive occurrence locations \ie $\mathbf{r}_i \sim\mathrm{Unif}( \{\mathbf{x}_1,\ldots,\mathbf{x}_N\})$.
This biases the training towards regions that contain valid data.




%
%
%
\subsection{Our Approach}
In this section we outline how we model and train our spatio-temporal prior $P(y|\mathbf{x})$.
\vspace{-5pt}

\vspace{-3pt}
\subsubsection*{Location and Object Embedding}
\vspace{-3pt}
In many contexts, different objects do not occur independently at a given spatio-temporal location.
Knowing that object A is present may provide information regarding the presence or absence of object B at the same place and time.
Similarly, different spatio-temporal locations are not independent, and may share commonalities.
We exploit this structure to encode low dimensional embeddings of objects and spatio-temporal locations.

Taking inspiration from \cite{chen2017DMSE}, we model our spatio-temporal prior as $P(y|\mathbf{x}) \propto s(f(\mathbf{x})\mathbf{O})$. 
Here, $f:\mathcal{R}^3 \to \mathcal{R}^D$ is a multi-layered fully-connected neural network that maps a spatio-temporal location $\mathbf{x}$ to a $D$ dimensional embedding vector.
$\mathbf{O} \in \mathcal{R}^{D\times{}C}$ represents an object embedding matrix, where each column is a different category.
The product $f(\mathbf{x})\mathbf{O}$ results in a $C$ dimensional vector, where each element represents the affinity that a spatio-temporal location $\mathbf{x}$ has for category $y$.
The intuition is that we are representing spatio-temporal locations and object categories in a shared embedding space where the inner product between the embedding of a location $\mathbf{x}$ and an object $y$ is large if $y$ is likely to occur at location $\mathbf{x}$. 
Finally, $s()$ is an entry-wise sigmoid operation to ensure that the resulting prediction are in the range $[0,1]$.

\vspace{-3pt}
\subsubsection*{Photographer Embedding}
\vspace{-3pt}
In online image collections we often have access to additional information at training time in the form of the photographer $p \in \mathcal{P}$ who captured the image.  
To see why this information is valuable, consider the following example. 
Suppose a photographer $p$ visits location $\mathbf{x}$ and does \emph{not} report object $y$. 
If $p$ has never taken an image of an object like $y$, then this non-report gives us little information.
However, if $p$ has a history of reporting categories similar to object $y$, then this constitutes weak evidence that $y$ might actually be absent at that location. 
Thus, we can interpret the same presence-only information in different ways depending on the individual who provides it.

To capture photographer biases, we embed photographers into the same shared embedding space as the objects and locations.
This is achieved by learning a photographer embedding matrix $\mathbf{P} \in \mathcal{R}^{D\times{}|\mathcal{P}|}$ at training time.
Like different object categories, photographers may have affinities for particular locations and times, and share similarities in their spatio-temporal patterns with other photographers.
This enables us to represent both a photographer's preference for a given location $P(p|\mathbf{x}) \propto s(f(\mathbf{x})\mathbf{P})$, and a photographer's affinity for a given object category $P(y|p) \propto s(\mathbf{O}^T\mathbf{P})$. 
Once trained, the photographer embeddings $\mathbf{P}$ are not required at test time, see Fig. \ref{fig:overview}.

\vspace{-3pt}
\subsubsection*{Joint Embedding Loss}
\vspace{-3pt}
Our goal at training time is to estimate the set of parameters $\theta = [\theta_f, \mathbf{O}, \mathbf{P}]$, where $\theta_f$ denotes the weights of the location embedding network $f()$, $\mathbf{O}$ is the category embedding matrix, and $\mathbf{P}$ is the photographer embedding matrix.

We start with the constraint that our model should be conservative \ie if a category $y$ has been observed at the spatio-temporal location $\mathbf{x}$ in the training set, then $s(f(\mathbf{x})\mathbf{O}_{:,y})$ should be close to 1, otherwise it should be close to 0. 
Here, $\mathbf{O}_{:,y}$ indicates the $y^{th}$ column of $\mathbf{O}$.
We rely on the location embedding function $f()$ to interpolate between presence locations.
This is conservative in the sense that it assumes that an object is absent if it has not been observed. 
This is a very strong assumption, but it enables the spatio-temporal prior to be aggressive in down-weighting incorrect predictions from the image classifier.

Our first loss encourages the model to predict the presence of objects where they have been observed in the training set and downweight their likelihood where they have not been observed: 
\begin{equation}
\begin{aligned}
\mathcal{L}_{o\_loc}(\mathbf{x}, \mathbf{r}, \mathbf{O}, y) =  & \lambda \log(s(f(\mathbf{x})\mathbf{O}_{:,y})) +  \\ 
& \sum_{\substack{i = 1 \\ i\neq y}}^C  \log(1 - s(f(\mathbf{x})\mathbf{O}_{:,i}))  + \\ 
& \sum_{\substack{i = 1}}^C  \log(1-s(f(\mathbf{r})\mathbf{O}_{:,i})).
\end{aligned}
\label{loss:pos}
\end{equation}
$\lambda$ is a hyperparameter used to weight the positive observations and $\mathbf{r}$ is a uniformly random spatio-temporal datapoint. 
Next, we want the affinity between a photographer $p$ and a location $\mathbf{x}$ be high if $p$ was present at $\mathbf{x}$, and low otherwise:
\begin{equation}
\begin{aligned}
\mathcal{L}_{p\_loc}(\mathbf{x}, \mathbf{r}, \mathbf{P}, p) = & \log(s(f(\mathbf{x})\mathbf{P}_{:,p})) + \\
 & \log(1 - s(f(\mathbf{r})\mathbf{P}_{:,p})).
\end{aligned}
\label{loss:ploc}
\end{equation}
We assume that a photographer has a low affinity for a category unless they have previously observed it: 
\begin{equation}
\begin{aligned}
\mathcal{L}_{p\_o}(\mathbf{O}, \mathbf{P}, y, p) =  & \lambda \log(s(\mathbf{O}_{:,y}^T\mathbf{P}_{:,p})) +  \\ 
& \sum_{\substack{i = 1 \\ i\neq y}}^C  \log(1 - s(\mathbf{O}_{:,i}^T\mathbf{P}_{:,p})).
\end{aligned}
\label{loss:pos}
\end{equation}


\noindent{}Finally, to estimate the parameters of our prior we maximize
\begin{equation}
\mathcal{L} = \mathcal{L}_{o\_loc} + \mathcal{L}_{p\_loc} + \mathcal{L}_{p\_o},
\label{loss:total}
\end{equation}
by iterating over each of the datapoints in the training set.

\begin{table*}[t]
\footnotesize
\centering
\begin{tabular}{| l | c |  c | c | c | c c c | c c c |} 
 \hline
  & {\bf YFCC} & {\bf BirdSnap} & {\bf BirdSnap}$\dagger$ & {\bf NABirds}$\dagger$ & \multicolumn{3}{c|}{{\bf iNat2017}} & \multicolumn{3}{c|}{{\bf iNat2018}}  \\ 
\hline
 $P(y|\mathbf{x})$ - Prior Type  & Test & Test & Test & Test & Val & Test Pu & Test Pr & Val & Test Pu & Test Pr \\ 
 \hline
 \hline
 No Prior (\ie uniform)                   & 50.15 & 70.07 & 70.07 & 76.08 & 63.27 & 64.16 & 63.63 & 60.20 & 50.17 & 50.33 \\
 Nearest Neighbor (num)     & {\bf 51.78} & 70.82 & 77.76 & 79.99 & 65.34 & 66.04 & 65.61 & 68.70 & 54.54 & 54.58 \\
 Nearest Neighbor (spatial) & 51.21 & 71.57 & 77.98 & 80.79 & 65.85 & 67.02 & 66.41 & 67.55 & 53.67 & 53.81 \\
 Discretized Grid           & 51.06 & 71.09 & 77.19 & 79.58 & 65.49 & 66.62 & 66.07 & 67.27 & 53.13 & 53.16 \\
 Adaptive Kernel \cite{berg2014birdsnap}       & 51.47 & 71.57 & 78.65 & 81.11 & 64.86 & 65.83 & 65.59 & 65.23 & 53.17 & 53.21 \\
 Tang \etal \cite{tang2015locationcontext}      & 50.43 & 70.16 & 72.33 & 77.34 & 66.15 & 67.08 & 66.53 & 65.61 & 54.12 & 54.25 \\
 {\bf Ours} no date           & 50.70 & 71.66 & 78.65 & 81.15 & 69.34 & 70.62 & 70.18 & 72.41 & 57.68 & 57.84 \\
 {\bf Ours} full            & - & {\bf 71.84} & {\bf 79.58} & {\bf 81.50} & {\bf 69.60} & {\bf 70.83} & {\bf 70.51} & {\bf 72.68} & {\bf 58.44} & {\bf 58.59} \\
 \hline
\end{tabular}
\caption{{\bf Classification accuracy.} Results after combining image classification predictions $P(y|I)$ with different spatio-temporal priors $P(y|\mathbf{x})$. All results are top 1 accuracy with classifier predictions extracted from an InceptionV3~\cite{szegedy2016rethinking} network fine-tuned on each of the respective datasets. $\dagger$ indicates that simulated locations, dates, and photographers from the eBird dataset \cite{sullivan2009ebird} are used. The baseline algorithms do not use date information.}
\label{tab:main_results}
\end{table*}

\begin{table}[h]
\footnotesize
\centering
\begin{tabular}{| l | c c c |} 
\hline
  & Top1  & Top3  & Top5 \\ 
 \hline
 \multicolumn{4}{|l|}{{\bf iNat2017 - InceptionV3 $299\times299$ }} \\
 \hline
  No Prior (\ie uniform) & 63.27 & 79.82 & 84.51 \\
  Ours no wrap encode & 69.48 & 84.43 & 88.15 \\
  Ours no photographer & 69.39 & 83.97 & 87.71 \\ 
  Ours no date & 69.34 & 84.16 & 87.89 \\ 
  Ours full & 69.60 & 84.41 & 88.07 \\
 \hline
 \multicolumn{4}{|l|}{{\bf iNat2018 - InceptionV3 $299\times299$ }} \\
 \hline
  No Prior (\ie uniform) & 60.20 & 77.90 & 83.29 \\ 
  Ours no wrap encode & 72.12 & 87.00 & 90.52 \\ 
  Ours no photographer & 72.84 & 87.30 & 90.75 \\
  Ours no date & 72.41 & 87.19 & 90.60 \\ 
  Ours full & 72.68 & 87.26 & 90.79\\
 \hline
 \multicolumn{4}{|l|}{{\bf iNat2018 - InceptionV3 $520\times520$ }} \\
 \hline
  No Prior (\ie uniform) & 66.18 & 83.32 & 88.04 \\ 
  Ours no wrap encode & 77.09 & 90.68 & 93.54 \\ 
  Ours no photographer & 77.64 & 90.82 & 93.52 \\ 
  Ours no date & 77.41 & 90.80 & 93.58 \\ 
  Ours full & 77.49 & 90.85 & 93.57 \\
 \hline

\end{tabular}
\caption{{\bf Ablation.} Classification accuracy for different variants of our prior on the iNat2017 and iNat2018 \cite{van2018inaturalist} validation sets. In the case of iNat2018, we still observe improvements when combining our prior with a more powerful image classifier - see rows `InceptionV3 $520\times520$'.}
\label{tab:ablation_results}
\vspace{-8pt}
\end{table}

%
%
%
\vspace{-5pt}
\section{Experiments}
\vspace{-5pt}
We evaluate the effectiveness of our spatio-temporal prior by performing experiments on several image classification datasets that have location and time information. 
We choose image classification because for other domains (\eg species distribution modeling) it is challenging to obtain accurate ground truth information regarding the true spatio-temporal distributions of the categories of interest. 

\vspace{-5pt}
\subsection{Datasets}
\vspace{-5pt}
While location metadata is readily available for online image collections, many popular image classification datasets do not contain this information \eg \cite{wah2011caltech,van2015building,deng2009imagenet,lin2014microsoft}.
Some datasets exist with location information, but for only a subset of the images \eg\cite{goeau2016plant}.
However, datasets containing images of different species of plants and animals are available with location, time, and photographer information.  
To this end, we perform experiments on the iNaturalist 2017 and 2018 (iNat2017 and iNat2018) species classification datasets which contain images collected and annotated by citizen scientists \cite{van2018inaturalist}.
They have 5,089 and 8,142 categories respectively. 
While \cite{berg2014birdsnap} evaluated their location prior on the BirdSnap dataset, the images and location metadata used are not provided by the authors. 
We recollect the images and location data from the web using the original image URLs. 
Despite the dataset consisting of images of species commonly found in North America, when we recollected the images and locations we found that the original images are from all over the world and 40\% were missing location.
Like \cite{berg2014birdsnap}, we also simulate location metadata for BirdSnap \cite{berg2014birdsnap} and another fine-grained dataset of birds, NABirds \cite{van2015building}, by associating each image with a species observation from eBird \cite{sullivan2009ebird}.
Our train locations and photographers are sampled from eBird 2015, and the test set is from 2016.
BirdSnap and NABirds contain images from 500 and 555 different species of North America birds.
Finally, we also perform experiments on YFCC100M-GEO100 \cite{tang2015locationcontext} (YFCC). 
YFCC contains 100 everyday object categories with associated locations, but no date or photographer information is provided.
The train and test split used in  \cite{tang2015locationcontext} is not available and so we created a new one.
Unlike the other datasets, many of the object categories in YFCC are not geographically distinct \eg `band', `ford', or `ipod'.

\vspace{-5pt}
\subsection{Implementation Details}
\vspace{-5pt}
Our location encoder $f()$ is a fully-connected neural network consisting of an input layer, followed by multiple residual layers \cite{he2016deep}, and a final output embedding layer. 
We jointly train the location encoder, along with the photographer and object embeddings using Adam \cite{kingma2014adam} for 30 epochs with a batch size of 1024, using dropout to prevent overfitting.  
The dimensionality of the shared embedding space is set to $D=256$.
When weighting the positive instances during training we set $\lambda$ to the number of categories. 
To counteract the heavily imbalanced nature of many of the datasets, we limit the maximum number of datapoints for each category per epoch. 
We set the maximum number of datapoints to 100, and for each epoch we randomly select a different subset for each category. 
The only exception is for YFCC, where capping the data hurt performance.
Details of our network architecture are in the supplementary material. 

Except where noted, at test time, our model takes three inputs -- longitude, latitude, and day of the year, specifying where and when the image of interest was captured. 
For these three input features $\mathbf{x}$ we explored different methods for `wrapping' the coordinates \ie an observation taken on December $31^{st}$ should result in a similar embedding to one captured on January $1^{st}$.
Similarly, we want geographical coordinates to wrap around the earth. 
To achieve this, for each input dimension $l$ of $\mathbf{x}$ we perform the mapping $[\sin(\pi x^l), \cos(\pi x^l)]$, resulting in two numbers for each dimension.
Here, we assume that each dimension of the input has been normalized to the range $x^l \in [-1,1]$.

For the image classifiers $P(y|I)$ we fine-tune a separate InceptionV3~\cite{szegedy2016rethinking} network for each of the datasets beginning with ImageNet initialized weights \cite{deng2009imagenet} with an image resolution of $299\times299$ (unless otherwise noted).

\subsection{Quantitative Evaluation}
\vspace{-5pt}
In Table \ref{tab:main_results} we evaluate how much our spatio-temporal prior improves image classification performance by comparing it to several baselines. 
We found that adding a uniform prior to the outputs of the nearest neighbor based baselines increases their performance.
This adds robustness in cases where there are no objects from the training set present near the test locations. 
The lack of this uniform prior explains the poor results for nearest neighbor based approaches in \cite{tang2015locationcontext}.
For the comparison to Tang \etal \cite{tang2015locationcontext}, we jointly train a linear layer to embed the raw location information along with an output layer to combine the location embedding with the features from the last linear layer of the image classifier. 
The rest of the weights of the image classifier are not updated.  
For each of the baseline algorithms we select their hyperparameters (\eg the number of neighbors) on a held out validation set for each dataset.
When location information is not available at test time, we assume a uniform prior over the categories. 

Our model performs on par, or better, than the baselines across all datasets. 
The advantage of our approach is that it is computationally efficient at test time and does not require features from the image classifier during training.
Compared to nearest neighbor based methods, it only requires a forward pass through a compact fully-connected neural network.
In addition, it also captures structural information such as object and photographer biases.
One failure case that is worth noting are the results on YFCC \cite{tang2015locationcontext}.
We observe that all methods perform similar to using no location information (No Prior).
This can be explained by the relative lack of spatio-temporal structure in the object categories present in the dataset.
Again, this is consistent with the findings in \cite{tang2015locationcontext}, where the authors had to use additional features to increase the performance.

\begin{figure*}[t]
    \centering
    \begin{subfigure}[b]{0.35\textwidth}
        \includegraphics[width=\textwidth]{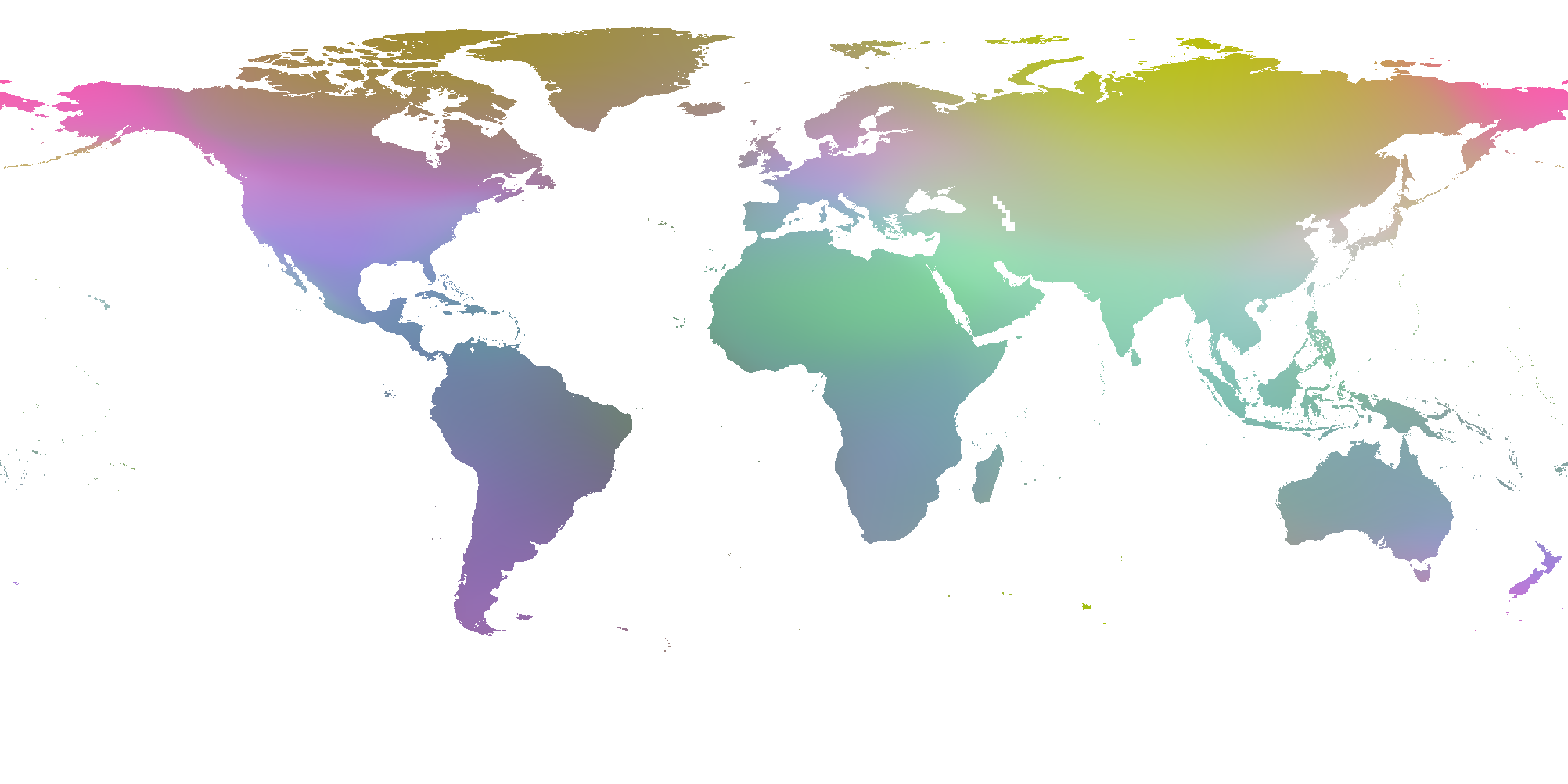}
        \vspace{-25pt}
        \caption{Location embedding}
        \label{fig:loc_embed}
    \end{subfigure}
    \hspace{50pt}
    \begin{subfigure}[b]{0.35\textwidth}
        \includegraphics[width=\textwidth]{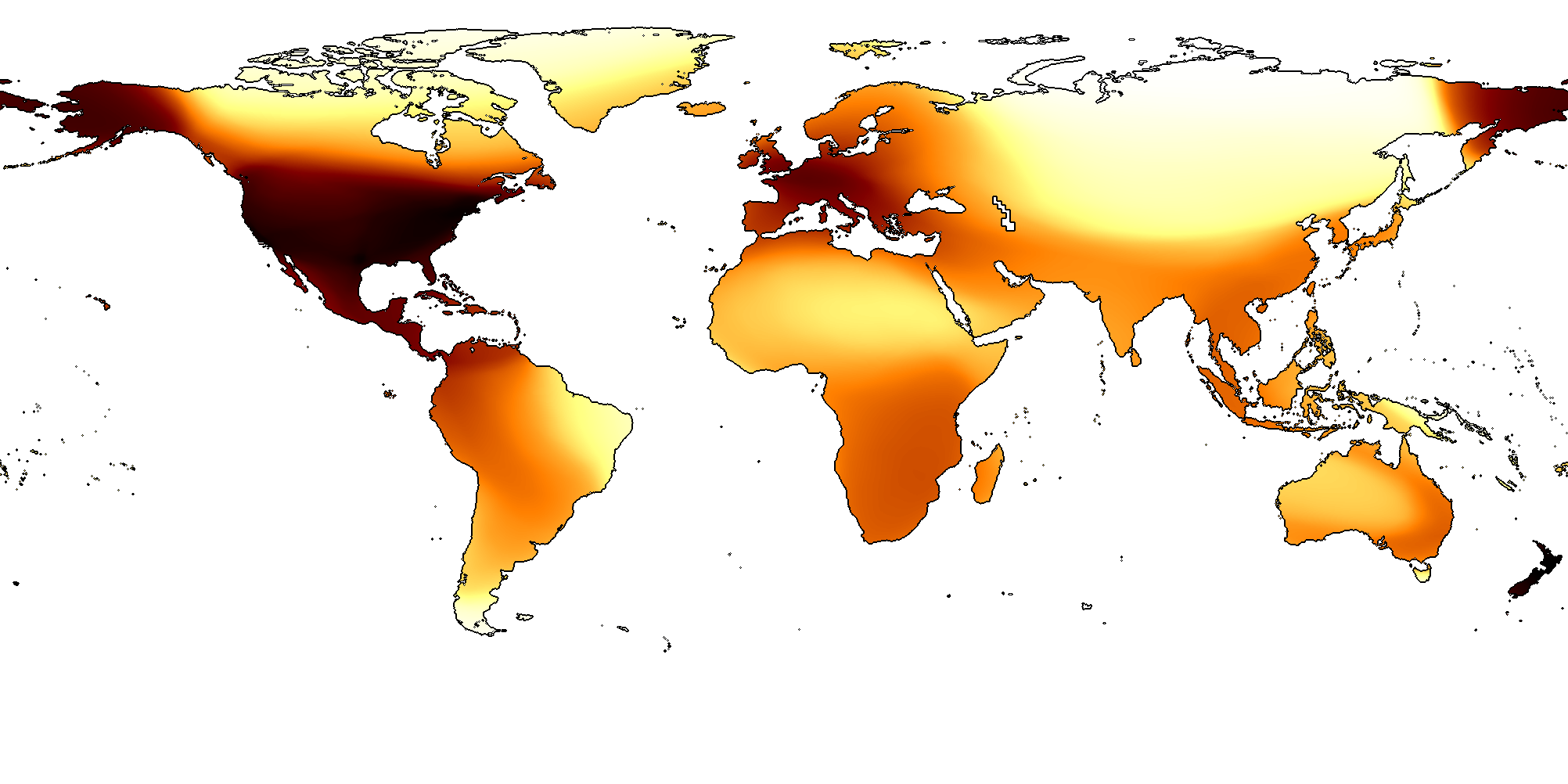}
        \vspace{-25pt}
        \caption{Photographer location affinity}
        \label{fig:user_loc_embed}
    \end{subfigure}
    \vspace{-5pt}
    \caption{{\bf Spatial predictions.} (a) Embeddings for each location on the earth for a model trained on iNat2018 \cite{van2018inaturalist}. We observe that the embeddings appears to capture information related to climate zones, despite not being trained on any climate data.
    (b) Log plot of estimated photographer location preferences. Darker colors indicate that more photographers have captured images in those locations. We can see that there is a large bias towards North America,  Europe, and New Zealand.}
    \label{fig:maps}
\end{figure*}

\begin{figure}[h]
\begin{center}
  \includegraphics[width=0.9\columnwidth,trim={0 0 0 0},clip]{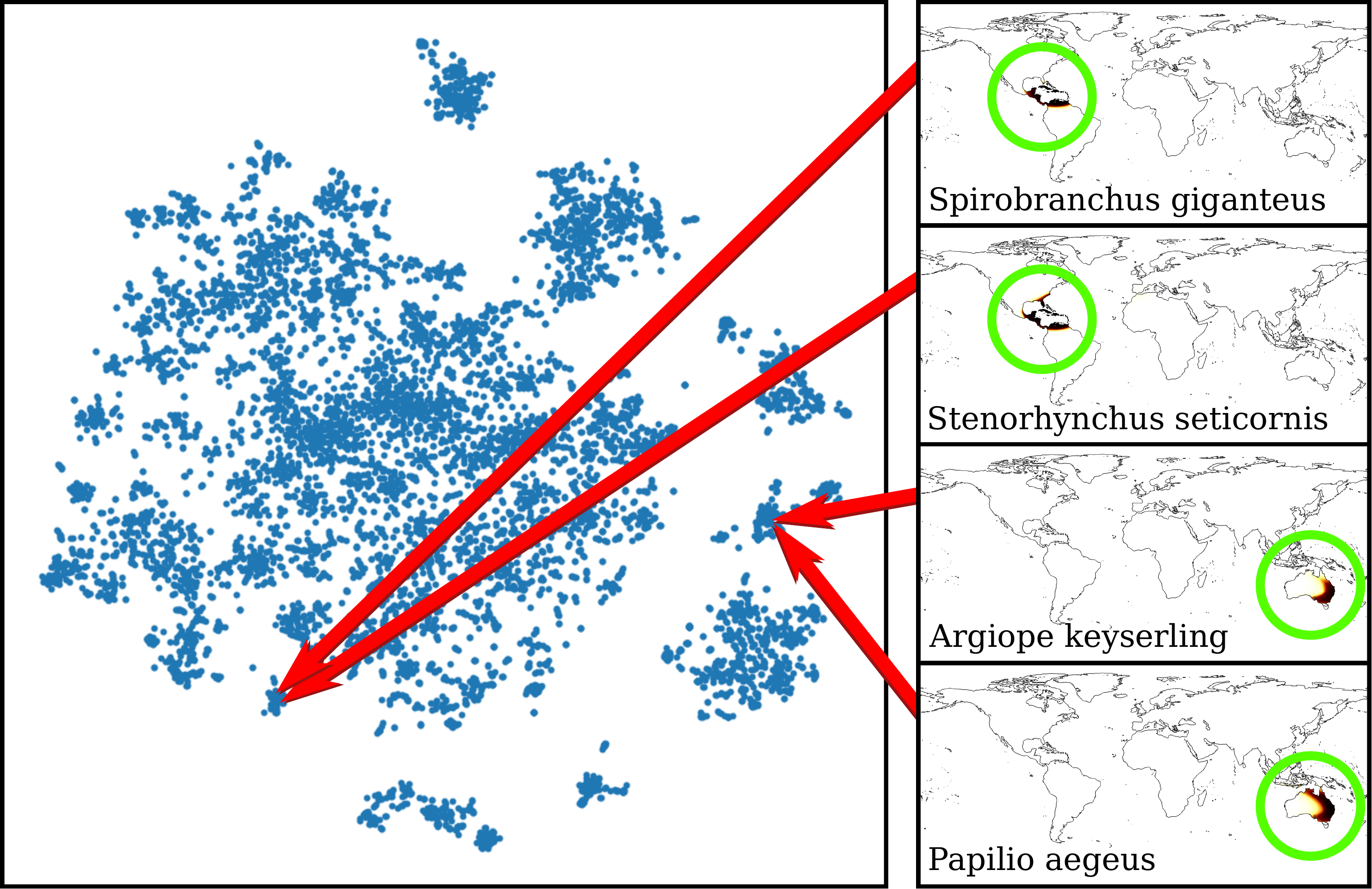}
\end{center}
  \vspace{-10pt}
  \caption{{\bf Object embedding.} t-SNE \cite{maaten2008visualizing} plot of the learned embedding $\mathbf{O}$ for all 8,142 categories from iNat2018~\cite{van2018inaturalist}. The location in the object embedding space encodes a category's preferences for a particular geographical region. We observe that categories that have similar spatio-temporal distributions tend to be close.}
\label{fig:class_embed}
\end{figure}

\vspace{-5pt}
\subsubsection{Ablation Study}
\vspace{-5pt}
In Table \ref{tab:ablation_results} we compare the performance of different variants of our model on iNat2017 and iNat2018~\cite{van2018inaturalist}.
Again, across all metrics there is a large increase in performance compared to the baseline uniform prior. 
In some cases, we even observe that there is an additional boost in performance when we explicitly model photographer biases. 

Training fine-grained image classifiers with larger input images can significantly increase classification performance~\cite{cui2018large}.
We observe that the benefit of our spatio-temporal prior is still apparent even when we use a more powerful classifier that has been training for longer with larger images. 
This increase in accuracy is also present when we evaluate performance using more lenient evaluation metrics \ie top 5 vs. top 1 accuracy. 
This is significant because it highlights that for some datasets the performance boost provided by the spatio-temporal prior is orthogonal to improvements in the underlying image classifier.

\begin{figure*}[h]
\begin{center}
  \includegraphics[width=0.95\textwidth,trim={0 0 0 0},clip]{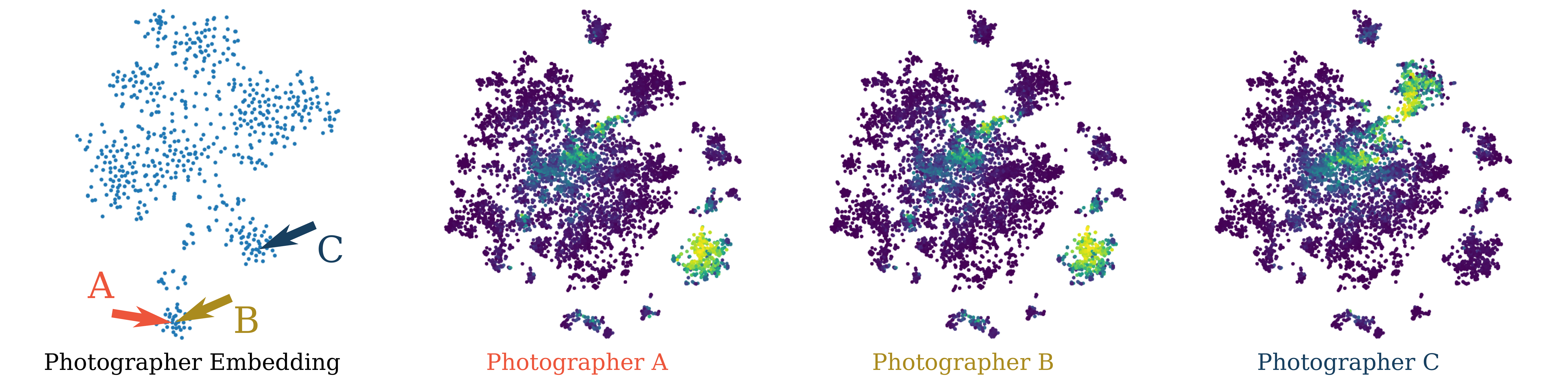}
\end{center}
  \vspace{-10pt}
  \caption{{\bf Photographer object affinity}. On the left we see a t-SNE \cite{maaten2008visualizing} plot of the photographer embeddings $\mathbf{P}$ for iNat2018~\cite{van2018inaturalist}. 
  The three plots on the right depict the predicted affinities for three different photographers (A, B, and C) visualized on the category embedding from Fig.~\ref{fig:class_embed}.
  Brighter colors indicate a higher affinity for a given category. 
  We observe that individuals that are close in the photographer embedding space $\mathbf{P}$ (\eg A and B) have similar category affinities, compared to those that are far away (\eg C).}
\label{fig:user_embed}
\end{figure*}

\begin{figure*}[h]
\begin{center}
  \includegraphics[width=0.98\textwidth,trim={0 0 0 0},clip]{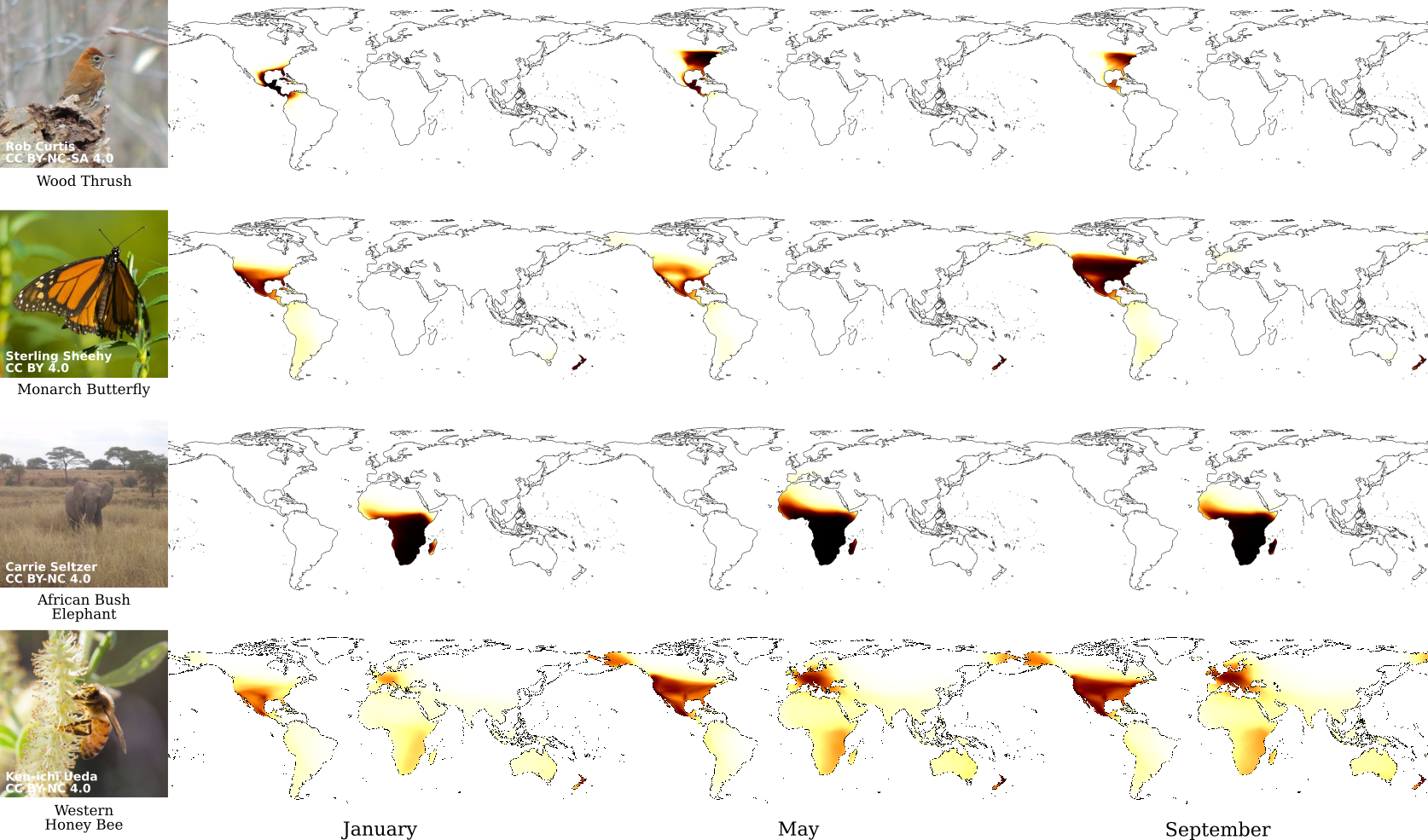}
\end{center}
  \vspace{-10pt}
  \caption{{\bf Spatio-temporal predictions}. Predicted distributions for several object categories for three different time points using our full model trained on iNat2018~\cite{van2018inaturalist}.
  Darker colors indicate locations where the categories are predicted to be found. 
  In the first two rows we observe that our model captures seasonal migratory behaviors. 
   On the bottom row, our model correctly predicts that the Western Honey Bee can be found on several different continents. 
  It is worth noting that the results are affected by geographical sampling biases in the iNat2018 dataset.}
\label{fig:temporal_evolution}
\end{figure*}

\subsection{Qualitative Evaluation}
Our model captures the relationship between objects, locations, and photographers.
In Fig. \ref{fig:maps} (a) we can see the resulting embeddings for each input location from our model trained on iNat2018 \cite{van2018inaturalist}. 
By applying the embedding function $f()$ to each location we can generate its $D$ dimensional embedding vector.
We then use ICA \cite{hyvarinen2000independent} to project the embedded features to a three dimensional space and mask out the ocean for visualization. 
Perhaps as expected, there is low frequency structure in the resulting image \ie nearby locations tend to support similar objects.
One advantage of our approach is that we are not restricted to a fixed discretization. 
As a result we can generate embeddings for any location and time.
In Fig. \ref{fig:class_embed} we visualize our learned object embedding $\mathbf{O}$.
Objects that have similar spatio-temporal distributions tend to result in similar embedding vectors. 

Distinct from other work, our prior also models the relationship between photographers and locations, and photographers and object categories.
In Fig. \ref{fig:maps} (b) we plot the estimated affinity for each input location across all photographers \ie $\sum_{p}s(f(\mathbf{x})\mathbf{P}_{:,p})$.
We only show results for photographers who provided at least 100 observations in the iNat2018 \cite{van2018inaturalist} training set, resulting in 634 individuals. 
In Fig. \ref{fig:user_embed} we display the estimated affinity for each object category for a set of photographers \ie $P(y|p) \propto s(\mathbf{O}^T\mathbf{P})$.
We observe that the embedding captures the similarity in object affinity held by different photographers.

Finally, in Fig. \ref{fig:temporal_evolution} we use our prior to generate spatio-temporal predictions for several different species from iNat2018~\cite{van2018inaturalist}.
Each image is generated by querying every location on the surface of the earth, on a specified day of the year, to generate $P(y=y^*|\mathbf{x})$ for the category of interest.
In practice, we evaluate $1000\times2000$ spatial locations for each time point (\eg first day of the month).
This step is very efficient as we can pre-compute $f(\mathbf{x})$ for every location, independent of the category of interest.
Again, for visualization we mask out the predictions over the ocean.

%
%
\subsection{Limitations}
We are limited by the quality of the provided location data \eg it can be inaccurate or intentionally obfuscated.
We also make strong assumptions about a photographer's affinity for an individual object category. 
In reality, these interactions may be complex \ie once a photographer captures an image of a particular category they may be less likely to take an image of the same object in the near future.
There are also known spatial biases in the types of citizen science data we use  \cite{beck2014spatial,chen2018bias}. 
However, this may not be a major issue as we can assume that the distribution of test locations and dates is similarly biased.
We currently only use location, time, and photographer ID during training. 
In practice, additional data such as environmental variables may be a valuable signal for specific object categories~\cite{geolifeclef2018}.

\vspace{-5pt}
\section{Conclusion}
\vspace{-5pt}
We introduce a spatio-temporal prior to help disambiguate fine-grained categories resulting in improved test time image classification performance.
In addition to helping image classification, our model also naturally captures the relationships between locations and objects, objects and objects, photographers and objects, and photographers and locations in an interpretable manner. 
Importantly, our prior is efficient at test time, both in terms of model size and inference speed, and scales to large numbers of categories.

\noindent{\bf Acknowledgements} This work was supported by a Google Focused Research Award and an NSF Graduate Research Fellowship (Grant No. DGE‐1745301). We thank Grant Van Horn and Serge Belongie for helpful discussions, along with NVIDIA and AWS for their kind donations.

{\small
\bibliographystyle{ieee_fullname}
\bibliography{geo_priors}
}

%
%
\clearpage

\begin{appendices}
Here, we present additional analysis of the results and details of the model presented in the main paper.

\section{Supplementary Results}
\begin{figure}[h]
\begin{center}
  \includegraphics[width=\columnwidth]{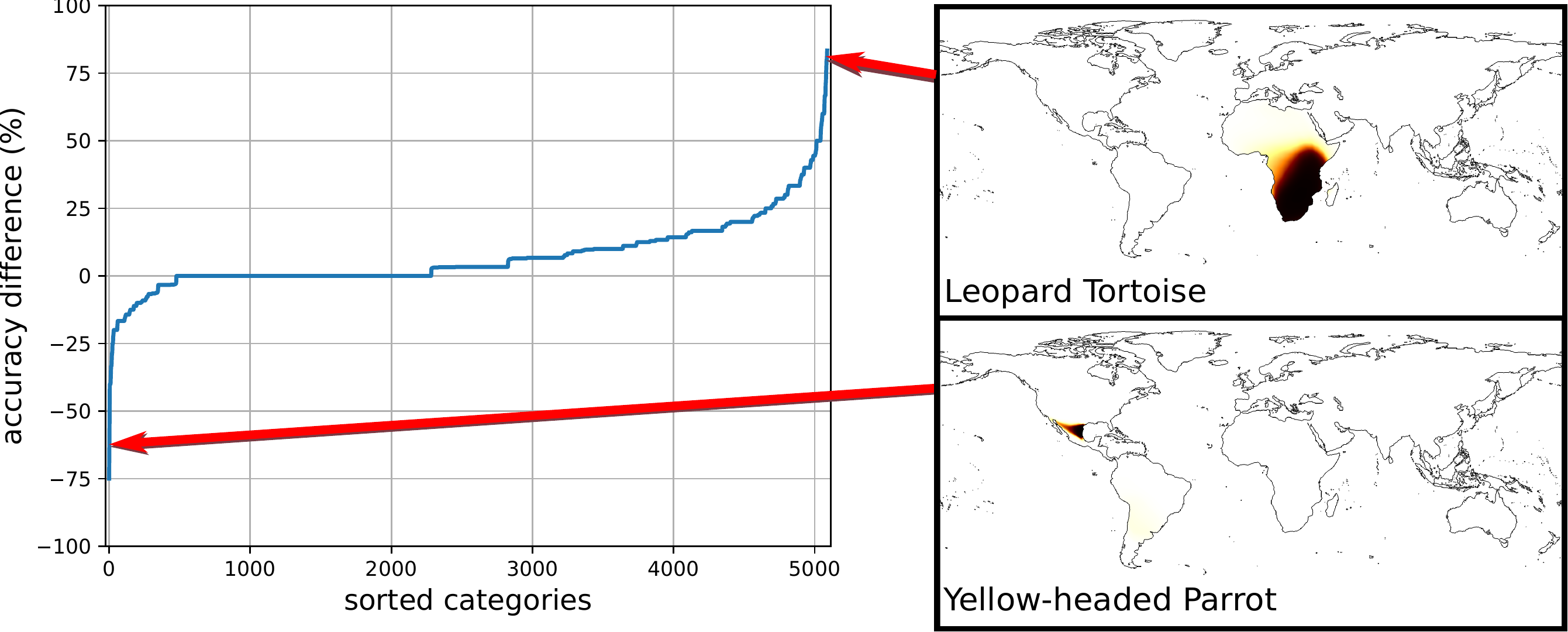}
\end{center}
  \vspace{-10pt}
  \caption{{\bf Accuracy improvement.} Here we see the categories in the iNat2017 validation split sorted by how much their classification accuracy improves after applying our spatio-temporal prior. 
  Values greater than 0 indicate categories where the spatio-temporal prior improves accuracy. 
}
\label{fig:improv}
\end{figure}

In Fig.~\ref{fig:improv} we observe the per-category accuracy improvement on iNat2017 \cite{van2018inaturalist} when using our spatio-temporal prior compared to no prior \ie the raw output of the classifier.
On the right of the plot we see large improvements for many categories \eg the `Leopard Tortoise' which is predicted to be found in east and south Africa.
Inspecting the errors before applying the spatio-temporal prior shows that the image classifier was confusing this category with the `Texas Tortoise', commonly found in Mexico and Texas.
Capturing this geographic specialization enables the prior to rule out instances of this category in other locations. 
On the left of the plot we see a small number of cases where the performance decreases after applying the prior \eg the `Yellow-headed Parrot'.
In this particular instance, the prior incorrectly biases the model by weighting the presence of the `Red-lored Parrot' more highly. 
The `Red-lored Parrot' has a very similar geographical range but with more observations in the training set (38 versus 18). 

\vspace{-5pt}
\section{Training Details}
\begin{figure}[h]
\begin{center}
  \includegraphics[width=\columnwidth]{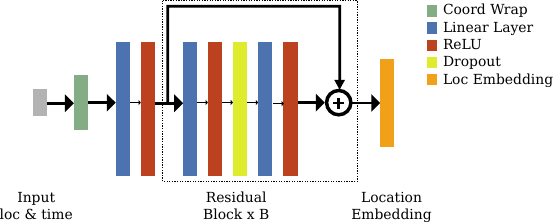}
\end{center}
  \vspace{-10pt}
  \caption{{\bf Location encoder.} Our location encoder $f()$ is a multi-layered fully connected neural network. }
\label{fig:network}
\end{figure}

In Fig.~\ref{fig:network} we illustrate the architecture of our location encoder $f()$.
As described in the main paper, each coordinate $x^l$ of the input spatio-temporal location vector $\mathbf{x}$ is mapped to $[\sin(\pi x^l), \cos(\pi x^l)]$, resulting in two numbers for each input dimension.
This is then passed through an initial fully connected layer, follower by a series of residual blocks, each consisting of two fully connected layers with a dropout layer in between. 
We set the number of hidden units in each fully connected layer and the output embedding to 256.
In total we use four residual blocks (\ie $B = 4$ in Fig.~\ref{fig:network}).

\section{Training Data Statistics}
\begin{figure}[h]
\begin{center}
  \includegraphics[width=\columnwidth]{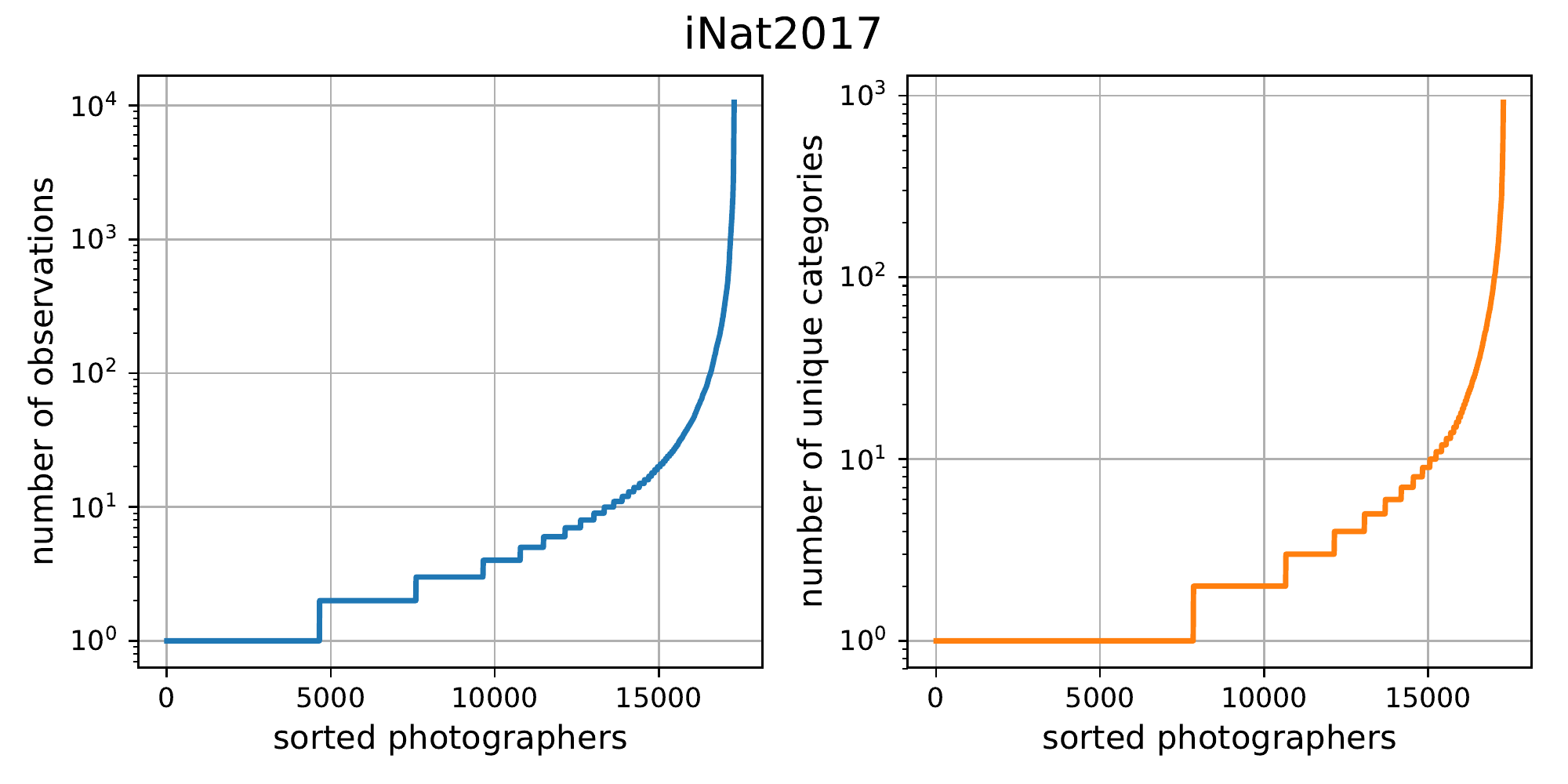}
  \includegraphics[width=\columnwidth]{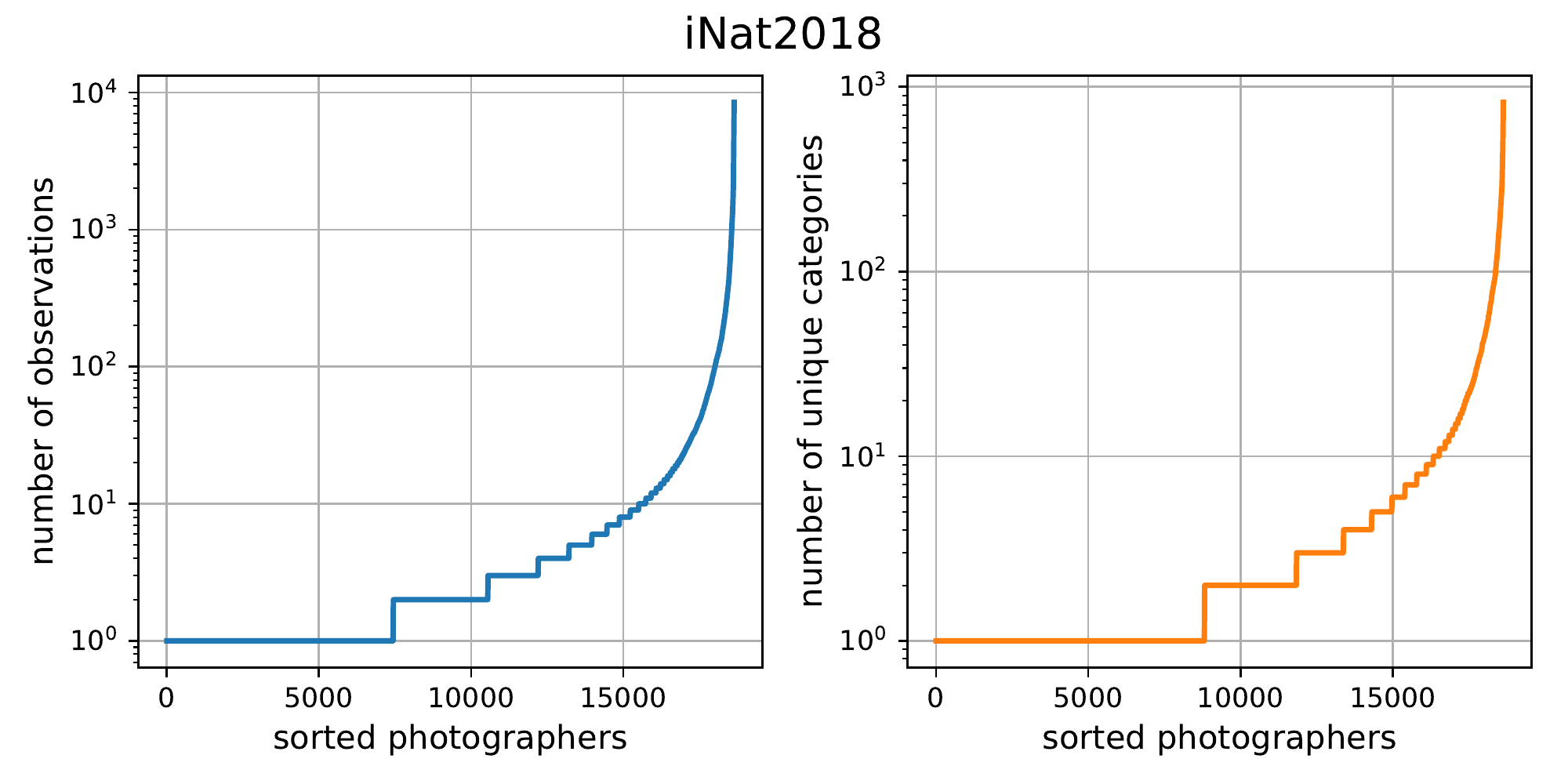}
\end{center}
  \vspace{-15pt}
  \caption{{\bf iNat training set statistics.} We see that for both iNat training sets most photographers only capture a small number of images. 
}
\label{fig:inat_md}
\end{figure}

In Fig. \ref{fig:inat_md} we display the total number of observations made by each photographer and the number of individual categories they observed for both iNat datasets. 
We only show data from the training set and exclude datapoints that do not have a corresponding valid location or photographer ID.
This results in 569,465 observations (\ie images) from 17,302 photographers for iNat2017 and 436,063 observations from 18,643 photographers for iNat2018. 

\end{appendices}
\end{document}